\def\year{2022}\relax
\documentclass[letterpaper]{article} 
\usepackage{aaai22}  
\usepackage{times}  
\usepackage{helvet}  
\usepackage{courier}  
\usepackage[hyphens]{url}  
\usepackage{graphicx} 
\usepackage{natbib}  
\usepackage{caption} 
\DeclareCaptionStyle{ruled}{labelfont=normalfont,labelsep=colon,strut=off} 
\usepackage{algorithm}
\usepackage{algorithmic}

\usepackage{newfloat}
\usepackage{listings}
\floatstyle{ruled}
\newfloat{listing}{tb}{lst}{}
\floatname{listing}{Listing}
\title{Mephisto: A Framework for \\ Portable, Reproducible, and Iterative Crowdsourcing}
\author{
    Jack Urbanek and Pratik Ringshia 
}
\affiliations{
    Meta AI\\
    jju@meta.com, tikir@meta.com
}

\usepackage{xcolor}

\begin{document}

\maketitle


\begin{abstract}
We introduce Mephisto, a framework to make crowdsourcing for research more reproducible, transparent, and collaborative. Mephisto provides abstractions that cover a broad set of task designs and data collection workflows, and provides a simple user experience to make best-practices easy defaults. In this whitepaper we discuss the current state of data collection and annotation in ML research, establish the motivation for building a shared framework to enable researchers to create and open-source data collection and annotation tools as part of their publication, and outline a set of suggested requirements for a system to facilitate these goals. We then step through our resolution in Mephisto, explaining the abstractions we use, our design decisions around the user experience, and share implementation details and where they align with the original motivations. We also discuss current limitations, as well as future work towards continuing to deliver on the framework's initial goals. Mephisto is available as an open source project\footnote{https://www.github.org/facebookresearch/mephisto}, and its documentation can be found at www.mephisto.ai.
\end{abstract}

\section{Motivation}
In the current era of Machine Learning (ML) research, we find a recurring gap between the discovery and widespread use of novel and best-practice methodology for high-quality crowdsourcing. This disconnect drives bad experiences both for researchers trying to collect high-quality data and workers who end up stuck with low-quality tasks and work practices. The thinking is, while some projects certainly aim to incorporate state of the art methodologies for crowdsourcing, those who do are left reproducing implementation work, and those who don't are left with worse results. It's impossible to directly confirm this claim, however studies show that even commonly used datasets have clear quality issues \cite{datadiscontent, testseterrors}.  

We believe much of this is caused by the common practice of publishing research papers without technical implementations of accompanying data collection methodologies. In a quick audit in November 2022 of PapersWithCode, a public repository of ML publications and datasets, we examine the top 35 cited datasets to see which ones provided code implementations. Of these, 31 describe their crowdsourcing methodology in varying levels of depth as part of the research, 18 papers were accompanied with \textit{usage} code at the time of writing, yet only 3 provide code for the \textit{collection and quality assurance} portion of the paper. If a researcher wanted to extend one of these existing common datasets, for instance for debiasing reasons, they would have to re-implement the collection or annotation flow from scratch.

This is in contrast to the modeling side of ML research, where code implementations have become much more standard. According to PapersWithCode's trends\footnote{https://paperswithcode.com/trends}, 8.5\% of papers in 2015 had code implementations compared to 27.75\% of papers from 2022. This increase closely follows the release and adoption of the TensorFlow and later PyTorch frameworks, and holds despite a huge increase in the raw count of ML and AI papers published every year. As 66-68\% of the repositories from 2022 are now built on PyTorch, it would be hard to conclude these frameworks had no role in driving increased code sharing.

Given the rise of data-driven ML solutions and adoption of ``big data'' more broadly \cite{mlinrealworld} there's even more pressure to improve the standards for data collection and annotation. Standard deep model architectures may be able to trade off between label accuracy and training time \cite{deeplyrobust}, but as we begin to run into scaling laws on large datasets \cite{scalinglaws} improving the quality of the dataset becomes a route for improving models again. While this is often done with pruning \cite{beyonddscalinglaws} it could also be handled at the start of the data funnel. Training time and cost are also becoming more central issues as large models can take months to train on energy-hungry hardware.

If data quality is a \textit{known} issue, we have to investigate why it isn't resolved. While annotation providers often try to build out tooling to improve labelling for common cases, these don't cover cases on the boundaries of research. Works that examine and analyze the issue, like \citet{datadiscontent}; \citet{betterusethecrowd}; and \citet{mlwcrowd}, stop understandably short of providing researchers with improved tooling. Further, change is difficult. It's not simple for most research labs to adopt better methodologies for running data collection, especially if they already have an established system or they cannot afford to invest the time and capital into building alternatives.

To resolve this need in the longer term, we aim to provide a route for efficient community collaboration for collecting high-quality data by using easily-to-publish shared tooling. This tooling is Mephisto. Making it easy to share, refine, and use best practices at the start of the data funnel can have impact on collection costs, training costs, fairness, and model accuracy.

The project is named Mephisto, alluding to the successor\footnote{https://en.wikipedia.org/wiki/Mephisto\_(automaton)} of the original ``man-in-the-machine'' chess playing `automaton' called the Mechanical Turk. As Mephisto improved upon the original ``man-in-the-machine'' operator with a remotely operated one, our framework aims to improve upon the complexities of traditional crowdsourcing, by abstracting its complexities further away and hopefully improving the experience for both workers and researchers in the process.

\section{Related Works}
Data annotation is certainly not an \textit{overlooked} area, however publications in the space of collection methodologies tend to fall roughly into a few categories. Specific area papers will devise a collection scheme for a specific task, and may explain the process. Surveys attempt to synthesize the learnings from these collections into broader themes. Tools and platforms attempt to solve a slice of the problem with clear code implementations.

\subsection{Novel Methodologies}

Across the field, there is no shortage of specific techniques that researchers have used to collect datasets. The following is entirely non-exhaustive, and a much more complete listing of these works can be found in \citet{betterusethecrowd}.

Task-specific interfaces have been developed for complex in-domain tasks \cite{labelMe, shapenet}.

Model-in-the-loop setups have been used for evaluation \cite{chatbotSafety}, sample-efficient data collection \cite{activelearning}, and live service \cite{scribe, hybridIntel}. Offline model evaluation is used as well \cite{whyTrust, acuteEval}.

Different techniques have been surfaced for ensuring data quality, expanding beyond inter-annotator agreement to more advanced approaches \cite{modthemod}. Some studies also specifically analysis the trade-off between data quality and budget \cite{accurateBudget}.

As seen in the top datasets though, publishing of these methodologies doesn't necessarily come with accompanying code (though some in the above do on Mephisto).

\subsection{Crowdsourcing Surveys}
To attempt to bring shared signal out of the spread of suggestions and methods, surveys of the field attempt to collect, group, and evaluate various techniques. These often provide suggestions for how others may include their own collections \cite{betterusethecrowd, mlwcrowd}. They understandably stop short of providing researchers with comprehensive tooling.

In many cases, surveys refer to crowdsourcing as a key part of the data lifecycle, and try to shed light on the complexities and next steps within a specific domain \cite{mllifecyle, thedatawork}. Some works have outlined entire workflows for developing higher-quality datasets \cite{crowdsourcingFramework}. While the insights are certainly valuable, few papers directly refer to these works when building out their tooling, and few of those end up releasing actionable code assets. 

One survey \cite{annotationsaurus} attempts to find and document all of the available annotation tools, though this list is certainly non-exhaustive.

\subsection{Full-code Solutions}
Often, works that release crowdsourcing code do so as part of a paper solving a contained problem space, like how \citet{legoeval} sets up compositional dialogue tasks, or making the experiment flow for research somewhat easier for a specific purpose, such as how \citet{psiturk} attempts to abstract the complexities of interacting with the Mechanical Turk platform.

Mephisto falls into this last category of work as well. We branch out of ParlAI-MTurk \cite{parlai}, a project that was designed to make dialogue-based research easier. Acknowledging the risk of becoming another bygone standard\footnote{For those familiar with the trend of https://xkcd.com/927/} we attempt to make the platform general enough to support integrating any of the above works, and intend to help ground the conversation in usable tools for researchers.

\section{Project Goals}

With Mephisto, we seek to address the core problems that prevent current crowdsourcing work from being easy to write, use, and distribute. For this we outline core values and elements we believe a research annotation platform should follow, such that we can evaluate our progress. 

\subsection{Distribution, Reproduction, and Extension}
In a research setting, each step of the process of distribution, reproduction, and extension are relevant for a work to contribute to forwarding the field. Distribution puts the techniques into the hands of other researchers, and can include sharing just methodology through the entire code setup. Reproduction allows new individuals to try out work, and the barrier for reproduction is often inversely proportional to how much of the process was able to be shared. Extension is the next step on reproduction, and pushes the initial work forward into new research.

To help facilitate distribution, reproduction, and extension of work, an annotation platform should make it easy to distribute all of the code related to a research project, and that code should be easy for new readers to set up, run, and modify on their own.

\subsection{Flexibility}
As research is a moving target, any platform that aims to support the varying needs of research should be casting a wide net for functionality. This complexity however runs counter to ease-of-use for a platform, which can raise the barrier of entry get researcher buy-in. An ideal platform should find a way to balance these two needs.
\subsubsection{Abstracting Crowdsourcing and Implementations}
In order to adapt to new scenarios and situations, it's valuable to examine the core elements of a crowdsourcing task, and isolate these into a coherent data model. From these building blocks multiple interoperable implementations can be built up, thus allowing for a high degree of control over specialized collection systems. For the average user, basic implementations with simpler controls can be provided to get them up-and-running with as little context as possible.

This type of approach allows work created on the platform to avoid the pitfalls of work such as in \cite{psiturk}, which helps users get started but ties them to a specific platform for crowdsourcing.
\subsubsection{Hooks with Default Best Practices}
Within individual components of the platform, it should be possible for researchers to exact a high degree of control to run specific jobs. This includes over portions such as worker-task pairing, collection pipelines and workflows, automated review tooling, and any other considerations. To prevent overwhelming new users, each of these should have best-practices provided by default, allowing new users to benefit from the shared knowledge of current best approaches.

This avoids the issues of both \citet{crowdsourcingFramework} and \citet{parlai}, where in the former there's too much flexibility at the onset (considering you would have to implement it all yourself), and in the latter all tasks are forced into Dialogue data collection best practices and techniques.

\subsection{Data Quality considerations}
Any project aiming to facilitate crowdsourcing must consider data quality to be a priority, as no matter how easy it is to use and share, it isn't particularly useful if the output data is low-quality. We raise a few important considerations in this space, many under the lens that research work is often time-limited and it can take a number of revisions and iterations to have something worthwhile.

\subsubsection{UI/UX and worker quality of life}
The designed user-interface (UI) of a task is a significant contributor to task result quality \cite{taskdesign}. Tasks that are well designed may contain clear criteria, include examples, give format specifications, reduce cognitive demand, etc. \cite{confusingthecrowd}. An ideal framework should help encode some of these best practices for the busy researcher.

Workers are also more likely to return to tasks that are designed with a good user experience (UX) in mind. Tasks should minimize user frustration, both in terms of design and usability. Tasks should also be architected so that they are resilient to errors. Hitting error cases should be clear to users at the least, and at most triggering some form of alerting so that the researcher can respond swiftly and appropriately.

Incorporating feedback channels is also a great way to identify and improve upon design blindspots that may occur. We consider these blindspots to be the norm, not the exception. An ideal crowdsourcing framework should provide researchers with feedback mechanisms that serve as a catch-all for any oversights on their parts. In implementation, this would allow for easy communication from workers back to the researchers through some feedback channel. Giving workers an opportunity to share feedback with researchers can create for a better worker experience. \cite{taskfeedback}.

\subsubsection{Quality Assurance Practices}
Standard quality assurance practices, such as worker qualification, gold-labelling, inter-annotator agreement, etc. should be easy to discover and enable without getting into the literature. Encouraging researchers to design their tasks with these elements in the forefront will result in better data quality than tasks with these elements added as afterthoughts. Further, common workflows like pilots and worker communication can be critical, and thus should be easy to enact.
\subsubsection{Worker Diversity and Representation}
When collecting a dataset, one element of quality comes from ensuring the data is worked on from as large and representative a collection of contributors as achievable. Often, this is limited by the tools of the company providing the crowd, and at times considered private information. Still, a strong crowdsourcing platform should provide tools to encourage researchers to extend their crowd with best-practices for simplified onboarding, task maximums per worker, and the ability to use multiple crowdsourcing platforms. At the very least, it should be able to report some metrics for the source crowd, possibly integrating with something like data cards \cite{datacards}.

\section{Current implementation}
In following alongside the values and principles from the previous section, we designed Mephisto with an underlying set of abstractions, a few initial implementations, and then some best-practice elements both for task quality as well as researcher experience. This section aims to give a technical overview of how Mephisto operates today.

\subsection{Abstractions}
We'll start off by describing Mephisto's underlying abstractions, which aim to break the complexity of crowdsourcing into components to build architecture around. After describing the data model, getting the rest of Mephisto is almost as easy as \textbf{A}rchitect, \textbf{B}lueprint, \textbf{C}rowdProvider, \textbf{D}atabase\footnote{Initialization sequence not by design, we promise}.
\subsubsection{What \textit{is} in a task? The Data Model}
In order to reason about crowdsourcing, we break out a number of definitions that represent underlying elements of the data model. 

The first is a \texttt{Task}, which can be considered as a group of directly related work that needs to be done, such as ``Label 50,000 images with varying segmentation masks''. 

Beneath this level is a \texttt{TaskRun}, which can be considered an individual job you may have run. Of the 50,000 images above, you may want to label the first 1,000 with 3 possible mask labels for a pilot. This would be an appropriate \texttt{TaskRun}. (A \texttt{Task} may have just one \texttt{TaskRun}, but will often have many).

Within a \texttt{TaskRun}, you may have many \texttt{Assignment}s, which can be considered a discrete element you need done. This starts to be at the level of what you'll show a worker, such as ``Label these 5 images with segmentation masks''. 

An \texttt{Assignment} may be broken up into many \texttt{Unit}s, which represent the contribution that one individual may have on a task. For some \texttt{Assignment}s, there may be just one \texttt{Unit}, however for example you may have two \texttt{Unit}s on an \texttt{Assignment} that you want to have labelled twice to check inter-annotator agreement, or on a dialogue where you need two workers to communicate with one another at the same time.

For those actually doing the work, we have \texttt{Worker}s which keep track of everything an individual has ever done for you for all tasks.

In order to distinguish the full \texttt{Worker} history, we also have \texttt{Agent}s, which can be considered as a pairing between a \texttt{Unit} and a \texttt{Worker} representing the work that worker did for that particular unit.

These are the underlying data model components that back Mephisto, and we can begin to reason about the rest of the flow for an annotation \texttt{Task} with this terminology.

\begin{figure}
    \centering
    \includegraphics[width=83mm]{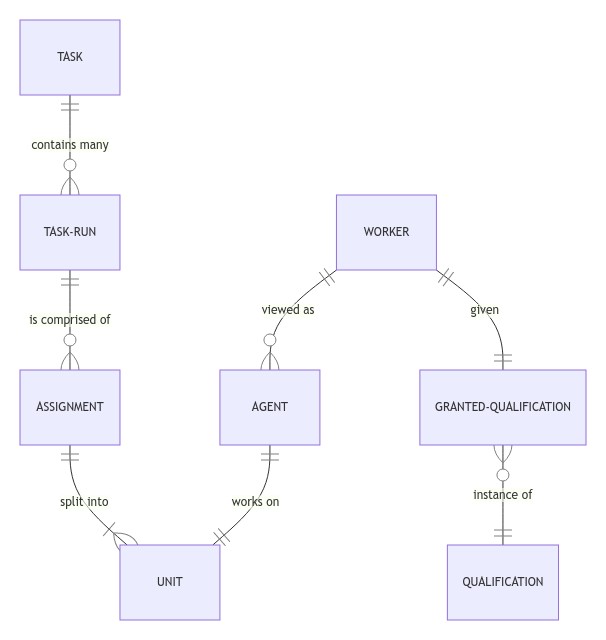}
    \caption{Mephisto Crowdsourcing Data Model Overview}
    \label{fig:data_model}
\end{figure}

\subsubsection{Hosting the Job: Architects}
\texttt{Architect}s comprise the scripts to set up a server that runs a task in Mephisto. They allow researchers to use Mephisto with different cloud configurations. For this, they cover the server lifecycle during a task, and thus should implement methods for preparing, deploying, and shutting down servers. They also define the interface for which external workers are able to connect to the Mephisto back-end.

\subsubsection{The Tasks: Blueprints}
\texttt{Blueprint}s are the center of Mephisto's different tasks, and aim to capture both task flows common to a task and configuration settings that can allow someone to customize and extend that task. They define the inputs and outputs for a specific task as well as the overall task interface. The specific abstraction requires a few important components, listed below:

\begin{itemize}
    \item The \texttt{AgentState} defines the format of the data that will be saved during collection of a \texttt{Unit}.
    \item The \texttt{TaskRunner} defines any back-end logic that is required to execute a task.
    \item The \texttt{TaskBuilder} defines any resources that need to be built before a job. Usually this includes the front-end to be hosted as part of a task.
    \item The \texttt{SharedTaskState} can be used to hold live state information shared between all of the \texttt{Unit}s in a \texttt{TaskRun}, often referred to when assigning work. 
\end{itemize}

\subsubsection{The Workers: Crowd Providers}
\texttt{CrowdProvider}s are what enable Mephisto to connect differing crowds to your task. These interact closely with the abstract \texttt{Worker}s, \texttt{Agent}s, and \texttt{Unit}s in the following way:
\begin{itemize}
    \item \texttt{<Crowd>Worker}s comprise the long-term identity for a worker, and are an interface where Mephisto can include worker-specific functionality that interfaces with a provider's API. This may include blocking, giving qualifications, and direct communication.
    \item \texttt{<Crowd>Unit}s are an interface to the remote hook of a job posting, or similar. They need to keep track of external status, and should also provide the interface for registering and expiring a work request with a provider.
    \item \texttt{<Crowd>Agent}s cover the link between a worker and a single \texttt{Unit}, and must implement methods for checking their remote status, as well as marking work as completed or rejected.
\end{itemize}

\subsubsection{Results Storage: Databases}
\texttt{Database}s are what enable Mephisto to store your results, regardless of the server setup you are using. For this, we provide the \texttt{MephistoDB} abstraction, which lists all of the required database calls one would need to implement to run Mephisto.

\subsection{Mephisto Architecture}
In practice, Mephisto is able to handle any arbitrary configuration of \texttt{Blueprint}, \texttt{Architect}, \texttt{CrowdProvider}, and \texttt{Database} and coordinate the initialization, deployment, monitoring, and shutdown of each over the \texttt{TaskRun} they comprise. Over the course of such a \texttt{LiveTaskRun}, it also reports metrics and saves partial results. One key goal is that any ``business logic'' that people would like to customize has a clear hook for doing so in the \textit{abstractions}, such that most users don't need to deal with the complexity of how these interfaces are coordinated. After data collection has concluded, Mephisto provides tools that allow one to interact with and explore the data stored in the \texttt{MephistoDB}.

\subsection{Blueprints}
It's our goal that in the majority of cases, most of Mephisto users should be able to rely on existing \texttt{Blueprint}s, rather than needing to write new ones from scratch. To this end, we provide a few useful implementations that cover a wide set of use cases.

\begin{itemize}
    \item The \texttt{ReactStaticBlueprint} is a setup where one can provide any simple data collection front-end application written in React that can be considered a single turn. In short: The worker is provided some data, they work on it remotely, and then return the result.
    \item The \texttt{RemoteProcedureBlueprint} allows a more complex setup, where the front-end application is able to make direct queries to some back-end specified during task setup. This allows for doing processing that wouldn't be possible on the worker's side, such as running a model in the loop.
    \item The \texttt{StaticHTMLBlueprint} stands as the easiest onboarding ramp to Mephisto, in that it accepts standard \texttt{.html} files that researchers may be more familiar with than React. It isn't as feature rich as other offerings though.
\end{itemize}

Beyond these, the \texttt{ParlAIChatBlueprint} stands as a good example of a live task with a specified and highly configurable flow, catered towards dialogue-focused jobs.

\subsubsection{Quality Assurance Mixins} We provide a handful of mixins for quality assurance which are available to be used on anything run from the \texttt{ReactStaticBlueprint} or the \texttt{RemoteProcedureBlueprint}. Including these mixins into a \texttt{Blueprint} means that blueprint has the specified functionality enabled and knows how to handle it.

\begin{itemize}
    \item The \texttt{OnboardingRequired} mixin allows researchers to set up a separate flow for workers who haven't done the task before, allowing them to learn what the requirements are.
    \item The \texttt{UseGoldUnit} mixin provides a familiar flow for providing known-good examples for which workers will be evaluated against periodically as a quality check.
    \item The \texttt{ScreenTaskRequired} mixin allows researchers to have the first actual \texttt{Unit} that a \texttt{Worker} works on for a job be a specified (usually easy-to-verify) unit, which allows researchers to do automated analysis and validation of before giving more work.
\end{itemize}
Researchers can use the primitives for these methods to incorporate strong quality assurance flows into their tasks, without needing to build any complex machinery on top of the underlying validation measures for their task. Of course, it still requires some initial rounds of piloting and tweaking to ensure the validators are well calibrated.

\subsection{Crowd Providers}
The main crowd providers we have implemented at the moment are the \texttt{MTurkProvider} and the \texttt{MockProvider}. The former allows for direct interfacing with the Amazon Mechanical Turk platform, while the latter allows for testing tasks locally, or allowing people to access while ``mocking'' a specified worker. Adding more Crowd Providers is ongoing work.
\subsection{Architects}
The currently available architects at the time of writing are the \texttt{HerokuArchitect} and the \texttt{LocalArchitect}. The former allows launching using Heroku cloud services as a provider. The latter allows hosting on the machine running Mephisto, which is useful for testing locally or collecting from research participants on the same local network. We also have an EC2-based architect which requires registering a domain name with AWS for use.
\subsection{Front-end Packages}
Another component of launching a crowdsourcing task is designing the task's UI. Mephisto allows researchers to launch tasks with UI implemented with either plain browser HTML, or for more advanced cases, with the React JavaScript library.

For the React implementation, Mephisto provides an npm (Node Package Manager) package named \texttt{mephisto-task}. The package enables researchers to interface seamlessly with the Mephisto back-end \texttt{Blueprint}s from their front-end code. It surfaces the task data provided from the back-end, callbacks to handle task submission for the \texttt{CrowdProvider}, as well as boolean flags that can be used to conditionally display different views (e.g. task preview, onboarding, errors, submissions, etc.).

The package also exposes three React Hooks: \texttt{useMephistoTask}, \texttt{useMephistoLiveTask}, and \texttt{useMephistoRemoteProcedureTask}. The latter two can be used for more advanced tasks, such as chat-bots or model-in-the-loop tasks, respectively. We particularly see model-in-the-loop as an opportunity to increase task result quality by augmenting worker performance in real-time, minimizing tedious and rote work, improving perceived UX, and providing real-time validation and feedback, though more research is needed in this area. Model-in-the-loop approaches can also be used to dynamically generate subsequent tasks based on prior tasks, providing  customized control over what tasks get launched next while a task run is already underway.

The \texttt{examples/} folder in the Mephisto GitHub repository provides sample task templates using the simple setup, as well as advanced setups including chat-bots and model-in-the-loop functionality.

\subsection{Worker Feedback}
With Mephisto's extensible architecture, creating plugins is easy as well. We provide two first-party plugins through the npm package \texttt{mephisto-worker-addons} to help improve the worker experience for tasks. Specifically, we provide the Feedback and Tips React components.

The Feedback component allows workers to provide suggestions back to the reseacher as they're working through tasks. This could include questions, bugs they've found, or positive acknowledgement. This communication channel back to the researcher can be a way to improve worker sentiment and improve task quality. Researchers can also choose to tip or give bonuses to submitters who provide valuable feedback. Communicating this reward scheme can also create a helpful incentive mechanism for gathering tips.

The Tips components allows workers to create a shared FAQ-style wiki that other workers can benefit from. This comes with built-in moderation as submitted Tips need to be approved by the researcher before they're visible to other workers. Aside from being helpful, these examples indicate a few ways of how Mephisto can be made extensible to suit custom research needs.

\subsection{Review Tooling}

Mephisto also includes a Python based command-line interface (CLI) tool to allow users to review a task's results, or more generally any arbitrary data. The command accepts an input data source as well as a ``review template'', and launches a local webserver to allow for browsing the data.

For any arbitrary data, one can just pipe in an input file:

\medskip
\noindent\texttt{cat input.jsonl | mephisto review --json my-review-interface --stdout}
\medskip

Or for using specifically with a Mephisto task run, one can use the \texttt{--db} flag:

\medskip
\noindent\texttt{mephisto review --db task-name my-review-interface --stdout}
\medskip

To facilitate review, we provide a React template based on \texttt{create-react-app} that implements a modular rendering architecture. This architecture allows researchers to easily define how a ``data item'' should be rendered by implementing their own custom renderer as a single React component. Out of the box, we ship a few default renderers; for example, a JSON renderer and a Word Cloud renderer for text-heavy tasks.

Once a task run is complete and a dataset has been accumulated, researchers can share results along with the Mephisto-based review and visualization tool as part of their publication. Mephisto's base review tooling was used by the Ego4D project \cite{ego4d} to share their collected 3,000 hours of egocentric video\footnote{https://ego4d-data.org/docs/viz/}.

\subsection{Worker Qualifications}
Mephisto provides a simple setup for tagging workers for any reason, wherein you can create and assign arbitrary \texttt{Qualification}s to any \texttt{Worker}. We find this is useful in setting up allow and block lists, querying or selecting workers based on skills you've noted them for, and creating complex task flows (such as those where participating in one role disqualifies another).

\section{Future work}
Mephisto is an evolving system, and we continue to iterate and develop it alongside the values listed in this document. As we discover new powerful methods for crowdsourcing we aim to include them in Mephisto as top-level functionality. We also aim to provide easy ways for anyone on the platform to build new hooks and functionality, and share them with others who are developing tasks. Further, we hope to extend the base set of existing tooling that Mephisto supports out-of-box. Lastly, we aim to extend the portability of the platform, such that it can be used with as many providers, on as many hosting solutions, and with as many tasks as possible. 

We also hope to continue to build along the dimensions of task-design - making it easier to share and use community-sourced design and task templates, opt into UI and UX best-practices as they emerge, and experiment with new primitives to improve worker experience, such as gamification.

Even with these steps though, we're only scratching the surface of implementing the best practices of today, let alone accommodating those of tomorrow. We hope this work can stand as a foundation that future work will build upon. Our roadmap is available on the Github project page, and we're open to feedback on where we should take the project.

\subsection{Contributing}
Mephisto is an open source project, and we value contributions from our users. We welcome anyone to join in and help with the vision of easy, reproducible crowdsourcing with best-practices built in on our GitHub\footnote{https://www.github.org/facebookresearch/mephisto}. Feel like we're doing something wrong, or are missing a technique that people should be using immediately? Great! File an issue, or better yet open a PR.

\section*{Ethical statement}
Mephisto is provided as a crowdsourcing software with a permissive license on use. While the Mephisto platform aims to improve annotation methodologies and facilitate cooperation towards resolving data collection issues, it certainly is still a work in progress towards those goals. It doesn't directly impose them as constraints on its users, so while we try to make currently agreed upon best practices the defaults, they can be overridden. 

As such, issues such as underpayment or mistreatment of workers, collection of biased datasets, and data licensing issues may still arise. A researcher using Mephisto still must to do their due diligence to ensure they are up-to-date on the best methodologies for their collection.

\section*{Acknowledgements}
We'd like to thank all of Mephisto's public contributors as well as the ParlAI team and other early pilot users.

\bibliography{aaai22.bib}
\end{document}